\documentclass[letterpaper, 10 pt, conference]{ieeeconf}  

\IEEEoverridecommandlockouts                              

\usepackage[bookmarks=true]{hyperref}
\usepackage{lipsum}
\usepackage{float}
\usepackage{placeins}
\usepackage{siunitx}
\usepackage{amsmath,amssymb,relsize,mathtools}
\usepackage{graphicx}
\usepackage{url}
\usepackage{todonotes}

\title{\LARGE \bf
Speeding up deep neural network-based planning of local \\ 
car maneuvers via efficient B-spline path construction
}

\author{Piotr Kicki and Piotr Skrzypczy{\'n}ski
\thanks{Authors are with Institute of Robotics and Machine Intelligence, Poznan University of Technology,
        ul. Piotrowo 3A, 60-965 Poznan, Poland
        {\tt\small name.surname@put.poznan.pl}}%
}

\begin{document}

\maketitle
\thispagestyle{empty}
\pagestyle{empty}

\begin{abstract}

This paper demonstrates how an efficient representation of the planned
path using B-splines, and a construction procedure that takes advantage
of the neural network's inductive bias, speed up both the inference and
training of a DNN-based motion planner.
We build upon our recent work on learning local car maneuvers from past
experience using a DNN architecture, introducing a novel B-spline
path construction method, making it possible to generate local maneuvers in
almost constant time of about 11 ms, respecting a number of constraints
imposed by the environment map and the kinematics of a car-like vehicle.
We evaluate thoroughly the new planner employing the recent Bench-MR framework 
to obtain quantitative results showing that our method outperforms state-of-the-art 
planners by a large margin in the considered task.


\end{abstract}



\section{INTRODUCTION}
Although autonomous vehicles are researched intensively, 
research on motion planning for these vehicles focuses mostly on managing traffic scenarios and rules \cite{urbandriving,highway}, paying less attention to the local maneuvers that are necessary to park a car in a crowded city center, to enter a shopping mall's garage, or to avoid a collision with another car that executes an unexpected maneuver.
Human drivers perform such local maneuvers intuitively, leveraging the experience from similar situations they have encountered in the past.
Unlike humans, planning algorithms still struggle to produce feasible paths in a very short time (usually fractions of seconds) avoiding collisions in dangerous situations, and satisfying the constraints of a car-like vehicle. 
A car is nonholonomic, has a limited steering angle and some physical dimensions, while the planned path should allow control with continuous velocity and acceleration due to safety and comfort considerations.
Although these requirements call for a solution that is rather a reactive behavior than a classic planning algorithm, reactive methods \cite{reactive} rarely produce smooth paths for the highly constrained 
scenarios we consider because of the vulnerability to local minima and the use of hard-coded heuristics.

\begin{figure}[tb]
    \centering
    \includegraphics[width=\linewidth]{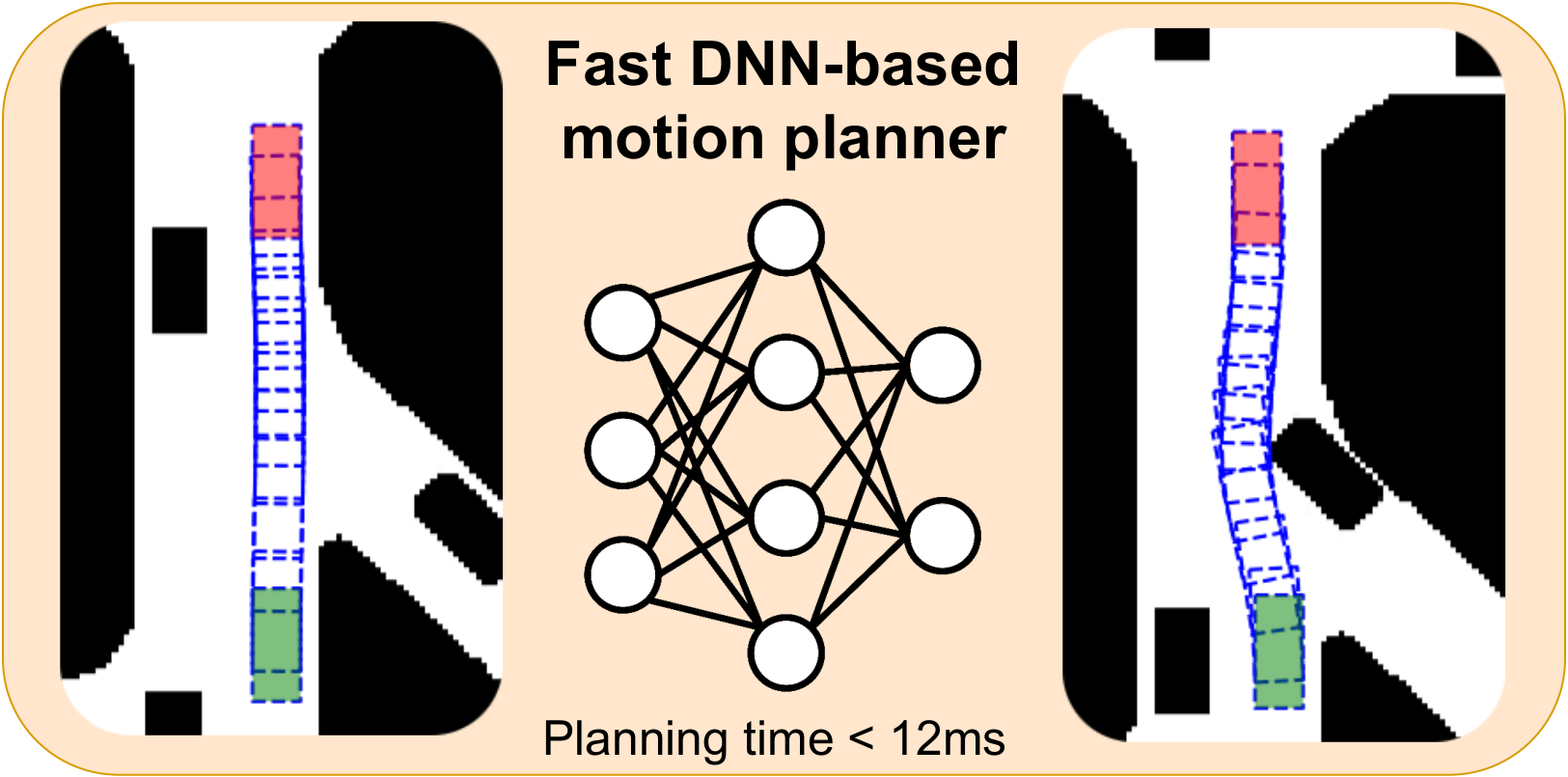}
    \vspace{-6mm}
    \caption{Motivational example of a traffic situation that requires rapid path planning in order to handle safely the unexpected maneuver of another road user (here the car coming from the right side)}
    \label{fig:main2}
      \vspace{-3mm}
\end{figure}

Here machine learning comes to the rescue, as modern methods, like deep neural networks (DNN), make it possible to learn even complicated decision-making policies in constrained state spaces \cite{deepreinforced}.
We have explored this idea in our recent paper \cite{eaai2021}, where we presented a neural network architecture and a training procedure that allow a local motion planner to learn from its own experience (Fig. \ref{fig:main2}). 
The neural network is trained with the use of a deep reinforcement learning framework, using gradient-based policy search \cite{policysurvey}. 
Learning through the interaction seems to carry out the most important information to improve the performance of the trained system \cite{harnessing}, while it does not impose any upper-bounds on it, unlike supervised learning, which performance is bounded by the quality of the reference trajectories or human demonstrations.


Although our previously introduced DNN~\cite{eaai2021} keeps the path computation time below $\SI{50}{\milli\second}$, some emergency maneuvers require even faster replanning, at the sensor frame rate of at least 30 fps, still yielding smooth paths satisfying all the constraints. 
Therefore, we contribute in this paper a novel path parametrization and procedure of its construction, which enables our method to compute yet better paths in an even shorter time in comparison to \cite{eaai2021}\footnote{Implementation of the proposed planner and code used for the experiments can be found at \url{https://github.com/pkicki/neural_path_planning/tree/bspline}.}.
The new parametrization, even though it follows the idea of the approximation of an implicitly defined oracle planning function, breaks up with the Markov Decision Process formalism used in \cite{eaai2021}, instead plans the whole maneuver at once.
The path is no longer represented with a sequence of polynomials, but with a single 7th degree B-spline curve.
The new representation and the matching architecture of the DNN guarantee that the planned path
reaches the goal configuration accurately, unlike the previous solution that usually produced small offsets.
The proposed path construction procedure introduces an inductive bias \cite{inductivebias} to the DNN,
which significantly speeds up the training.

We evaluate thoroughly the new planning method using the latest motion planning benchmarking
software Bench-MR \cite{benchmr}, which allows us to demonstrate that our solution is
superior to both state-of-the-art planners and the solution from \cite{eaai2021} in the considered task of local maneuver generation.

\section{RELATED WORK}
The problem of motion planning for vehicles has been tackled multiple times in the literature
with a broad spectrum of algorithms following different paradigms, such as graph search \cite{thrun},
sampling-based planning \cite{karaman}, trajectory optimization \cite{arab,kinodynamic} or probabilistic motion primitives \cite{prompt2021}.
For the last two decades, sampling-based algorithms are considered the group of
methods most suitable to efficiently plan paths in complicated configuration spaces.
Extensions of the Rapidly-exploring Random Trees (RRT) concept allow
asymptotically optimal planning \cite{karaman}, while more recent algorithms,
such as Batch Informed Trees (BIT$^*$) \cite{BIT} or Adaptively Informed Trees (AIT$^*$)
\cite{ait2020} quickly find a feasible path, and if the computational time
permits, converge towards the optimal one.
Unfortunately, few of the sampling-based algorithms intrinsically consider the constraints of a kinematic car \cite{rrtcar}.
Fast path planning respecting the non-holonomic constraints is possible using
for example the State Lattices algorithm \cite{SL}, but requires motion
primitives with 0 curvature at the boundaries to preserve the curvature continuity.
Using the continuous curvature Dubins curves \cite{ccdubins} (cc-Dubins) with some sampling-based
algorithm leads to the paths that achieve unnecessarily high curvatures.
Hence, the planning results, even if computed as quickly as from our DNN,
are not equivalent to our smooth paths that can be easily followed by a standard controller \cite{VFO}.

Only a few of the planning algorithms utilize the experience gained in the tasks executed before or some easily accessible {\em a priori} knowledge in order
to improve future planning attempts.
This concept is relatively new in robotics \cite{berenson,experience},
however in a way innate and natural for humans, as we all use the experience
gathered throughout life to improve our future actions.
Most of the works in this field consider ways to improve the speed of sampling-based
motion planners by biasing the sampling distribution \cite{ichter} or directly
predicting the next node \cite{mpnet}.
The concept of teaching neural networks how to plan has been applied in several recent papers \cite{harnessing,oracle2019,mpnettro}, however, all of those approaches have considered only a reaching problem for manipulators.
The use of deep reinforcement learning for motion planning in autonomous vehicles in a context
wider than path planning is surveyed in the recent paper \cite{titsurv}.
Among the works that leverage neural networks for vehicle path planning \cite{imitation} uses DNN and data collected from human drivers to learn a driving policy in urban scenarios, while \cite{shorterm} deals with local maneuvers with an end-to-end approach that casts the task of selecting the steering angle as a classification problem. 
Recently, Xiao {\em et al.} \cite{stone2021} introduced the novel Learning from Hallucination technique to address vehicle motion planning in highly constrained spaces.
A jointly learnable behavior and trajectory planner for self-driving vehicles was introduced in \cite{urtasun}.
Unlike the majority of neural planning methods that rely on paths demonstrated by experts, we apply a deep reinforcement learning approach \cite{deepreinforced}, thus conserving both the time and human effort in the training phase.

\section{NEURAL NETWORK LOCAL PATHS GENERATION}

\subsection{Problem definition}
In order to focus the presented research on a concrete problem of practical value, we define the task of planning a feasible monotonic, curvature-continuous path
$\mathcal{P}$ from an initial state $q_0$ to the desired state $q_d$, taking into account the geometry of the vehicle (modeled with a rectangle),
the local environment (a grid map of the size $\SI{25.6}{\metre} \times 
\SI{25.6}{\metre}$
represented with resolution $\SI{0.2}{\metre}$), and typical kinematic
constraints of an Ackermann steering car (no lateral and longitudinal slip, and limited steering angle $\beta$).

\subsection{Proposed solution}


We observe that different path planning methods, although they are algorithms,
can be viewed as mappings from some problem's space to the space of the
solution.
Thus, we can consider a special type of such a mapping, that for all elements of 
some set of motion planning problems returns a single feasible path that solves 
the given problem instance.
We have introduced functions which have this property as \textit{oracle planning 
functions} in \cite{eaai2021}.
It is hardly possible to define explicitly such a function for a broad set of practical planning problems.
However, knowing the properties of such a function we can define it implicitly for a specified problem, by introducing the constraints and quality criteria imposed on its output.
Because of this, we propose an oracle planning function approximator, which tries to mimic the oracle planning function behavior by being optimized to produce motion plans that satisfy the constraints and maximize the quality criteria.

\begin{figure*}[t!]
    \centering
    \includegraphics[width=0.7\linewidth]{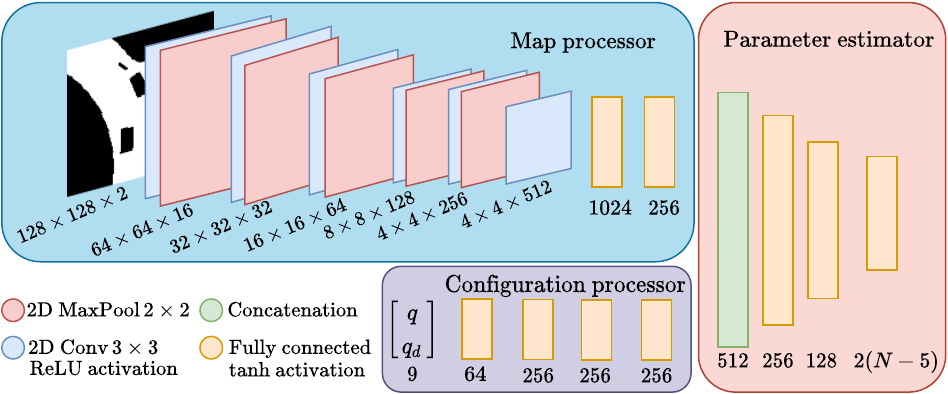}
    \caption{Architecture of the proposed neural network with the parameter estimator for B-spline-based paths}
    \label{fig:nn}
    \vspace{-1mm}
\end{figure*}

We propose to train a neural network how to plan feasible paths that solve the
instances of the introduced problem i.e. how to approximate the oracle planning function.
This neural network takes as input a map of the environment together with initial $q$ and desired $q_d$ vehicle configurations, processes these data, and returns a representation of the path.
We use a neural network which is a further development of the architecture
proposed in \cite{eaai2021}, but 
its final processing block (the parameters estimator) is changed, in order to take advantage of the novel representation of the generated path (see Fig.~\ref{fig:nn}) we introduce here. 
The new way the resultant path is constructed makes it possible to generate the final path from a single neural network prediction.
This, in turn, enables the new method to plan extremely fast and to produce 
paths that are smoother and have a warranty of reaching the goal.
Furthermore, the new architecture introduces some important inductive bias to the neural network, which speeds up its training significantly.

\subsection{Path representation}
In contrast to \cite{eaai2021}, we propose here to represent the solution path
with a 7th degree B-spline curve \cite{bspline}. This representation allows us to define a path, parametrized by its length, that starts and ends at given 
positions, with certain orientation and initial curvature, thus we can ensure 
that the boundary constraints imposed by the given planning problem are satisfied. 
The new B-spline representation is essential for allowing the neural network to 
generate the path in a single inference, making the planning process very fast.

\subsection{Neural network based path generation}
To define a 7th degree B-spline one has to provide control points $p_i = (x_i,
y_i) \in \mathbb{R}^2$ for $i=0,1,\ldots,N-1$. While 5 of them (the first 3 and the last 2) are defined already by the boundary conditions of the planning problem, the rest have to be determined by the neural network. 
Unfortunately, allowing the neural network to simply produce some control points on the map does not work. 
A randomly initialized neural network produces a bunch of control points randomly placed around the center of the map, which results in an extremely tangled curve that intersects with itself and is hard to entangle taking into account the curvature constraints.

Therefore, considering the task of planning local maneuvers for a car-like vehicle, we propose to organize the neural network predictions in a binary tree--like structure to produce paths without self-intersections. Each control point
$p_i$ is determined with the use of another two adjacent control points ($p_j, p_k$) that are already in the tree, 
such that a neural network prediction is scaled to fit in the square, with a 
side length of $d_{j,k} = \max\{|x_j - x_k|, |y_j - y_k|\}$, centered in the 
middle of those points. Thus, the resultant position of the control point $p_i$ can be defined by
\begin{equation}
    p_i = \frac{p_j + p_k}{2} + \frac{1}{2} d_{j, k} \cdot (\phi_{2(i-3)}, \phi_{2(i-3)+1}),
\end{equation}
where $(\phi_{2(i-3)}, \phi_{2(i-3)+1}) \in [-1; 1]^2$ are two predictions made by a neural network.

\begin{figure}[hbp!]
\vspace{-2mm}
    \centering
    \includegraphics[width=\linewidth]{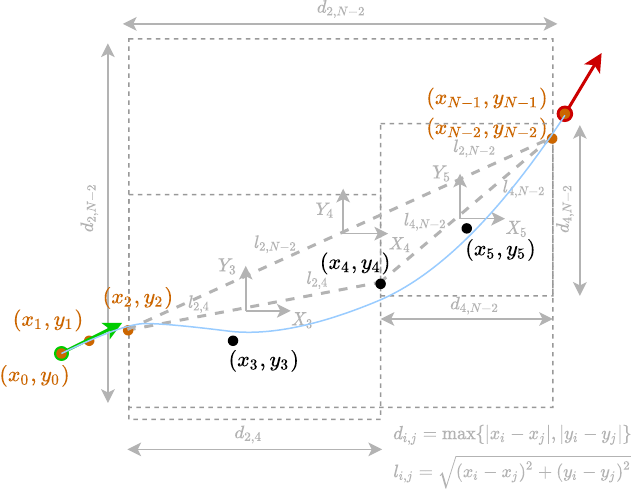}
\vspace{-6mm}    
    \caption{General idea of defining the B-spline control points. Brown control 
points are stemming from the motion planning problem boundary conditions, while 
black ones are determined by the neural network output and two adjacent control 
points which defines the area where the new point can be placed}
    \label{fig:tree} 
\end{figure}

The tree of control points, is built starting from the root $p_{2+2^{D-1}}$, where $D$ is the tree depth.
To define the root of the tree $p_{2+2^{D-1}}$, the $p_2$ and $p_{N-2}$ boundary control points, and two 
neural network predictions $(\phi_{2(2^{D-1}-1)}, \phi_{2(2^{D-1}-1)+1})$ are used (because of two-dimensional control points). 
Next, points at the second level $p_{2+2^{D-2}}, p_{2+2^{D-1}+2^{D-2}}$ are defined. They have to lie in between root and the boundary control points, therefore they are  determined using analogue elements of the neural 
network output and the pairs of control points $p_2, p_{2+2^{D-1}}$ and
$p_{2+2^{D-1}}, p_{N-2}$. 
Similarly, the next tree level can be populated with
another 4 control points, using corresponding neural network outputs and all points which are already in the tree, together with the aforementioned boundary control points $p_2$ and $p_{N-2}$.
In the same way, the next 8 points can be added in between the control points of the previous tree level.
The idea of the proposed path generation procedure, for $D=2$, is described graphically in Figure~\ref{fig:tree}.

Such a way of constructing the path from the neural network outputs is beneficial in terms of the feasibility of the training process and introduces an inductive bias into the inference and learning phase. 
Due to this representation, the neural network, even randomly initialized, favors paths that have smaller curvature, do not self-intersect, and are ensured to connect initial and goal configurations.
To fit this representation, we proposed the neural network architecture presented in Figure~\ref{fig:nn}. Its outputs are in the range $[-1; 1]$ due to the use of the $tanh$ function, and are interpreted as $x$ and $y$ coordinates of the control points in the coordinate systems determined by their parents in the tree.

\subsection{Differentiable loss function}
To train this neural network, we utilize the idea of the implicitly defined 
\textit{oracle planning function} and formulate a loss function:
\begin{equation}
\label{eq:loss}
  \mathcal{L} = \mathcal{L}_{curv} + \mathcal{L}_{coll} + 
\sigma_{tcurv}\gamma\mathcal{L}_{tcurv},
\end{equation}
where $\mathcal{L}_{curv}$ is a loss that penalizes paths that violate the 
curvature constraints, $\mathcal{L}_{coll}$ penalizes paths that are not 
collision-free, wheres the $\gamma\mathcal{L}_{tcurv}$ term minimizes the sum of changes in curvature along the path, but only if the indicator function $\sigma_{tcurv} = 1$, i.e. 
when other loss terms are zeroing out. Such a construction of the loss function 
stimulates the oracle planning function approximator to produce paths that have 
the same properties as the oracle planning function ($\mathcal{L}_{curv} = 
\mathcal{L}_{coll} = 0$), and are characterized by reduced changes in the steering angle $\beta$ along the path due to
the total curvature loss $\mathcal{L}_{tcurv}$ scaled with the parameter $\gamma$, which in our experiments was set to $0.1$. 

While all components of the loss function \eqref{eq:loss} are inspired by the ones introduced in \cite{eaai2021}, we redefined the way how the
$\mathcal{L}_{curv}$ and $\mathcal{L}_{tcurv}$ losses are calculated and used almost the same procedure for the collision loss $\mathcal{L}_{coll}$.

Due to the B-spline representation, we are able to easily obtain the first and second derivatives of the generated path using its control points, instead of relying on the local approximations, as it was done in \cite{eaai2021}.
Therefore, to calculate the curvature $\kappa$ of the path, we sample the 1024 equally distant points on the B-splines of its first and second derivatives, and obtain 1024 samples of the curvature $\kappa_0, \kappa_1, \ldots, \kappa_{1023}$. Using these samples we can define the curvature loss $\mathcal{L}_{curv}$ by
\begin{equation}
    \mathcal{L}_{curv} = \sum_{i=0}^{1023} \max\left(|\kappa_i|-\kappa_{max}, 0\right),
\end{equation}
where $\kappa_{max} = \frac{1}{R_{min}}$ is the maximum allowed path curvature, that stems form the vehicle minimal turning radius $R_{min}$.
Whereas, total curvature loss $\mathcal{L}_{tcurv}$ is defined by
\begin{equation}
    \mathcal{L}_{tcurv} = \sum_{i=1}^{1023} |\kappa_i-\kappa_{i-1}|,
\end{equation}
while collision loss $\mathcal{L}_{coll}$ is defined by
\begin{equation}
\label{eq:collision_loss}
 \mathcal{L}_{coll} = \sum_{i=1}^{1023}
 \sigma_{coll}(\Pi_{i}, \mathcal{F})
  \sum_{k=0}^{4} 
  d(\mathcal{P}_r, \Pi_{ik}) l_{i},
\end{equation}
where $\sigma_{coll}$ is the collision indicator function, which is equal to $1$ if any part of the vehicle circumference $\Pi_i$ at $i$-th point on a curve lies outside the free space $\mathcal{F}$, $d(\mathcal{P}_r, \Pi_{ik})$ is the Euclidean distance between the $k$-th characteristic point (4~corners of the rectangular body of the vehicle and the guiding point in the middle of the rear axle) and the reference path $\mathcal{P}_r$, whereas $l_{i}$ is the Euclidean distance between $i$-th and $(i-1)$-th points on a path.
Notice that the auxiliary reference path is used during learning only to evaluate the
$\mathcal{L}_{coll}$ component of the loss function, suggesting how to retreat from the collision areas.
As we do not evaluate the similarity between this path, and the learned one, the auxiliary path does not give an upper bound for the performance of our algorithm.
Whereas the reference path can be produced by any planning method for nonholonomic, car-like vehicles,
we use State Lattices \cite{SL}, mostly because of its simplicity and speed.





Importantly, all considered losses are differentiable with respect to the neural network outputs and thus the gradient of the proposed loss function can be used to directly optimize the neural network weights using gradient descend algorithms.

\section{EXPERIMENTAL EVALUATION}
To evaluate the proposed B-spline neural network planner, we trained it for 400 epochs with a learning rate $5\cdot10^{-4}$ and batch size $128$ on the training dataset introduced in \cite{eaai2021}, which consists of more than 130000 scenarios on 23517 local maps obtained from both simulated and real missions of autonomous cars.
We used the same model of a car as in \cite{eaai2021}, with maximal admissible curvature $\kappa_{max} = \SI{0.227}{\per\metre}$, length equal to $\SI{4.05}{\metre}$ and width of $\SI{1.72}{\metre}$. For the training NVIDIA GTX1080Ti GPU was used, while for inference an IntelCore i7-9750H CPU was sufficient to obtain the required planning times.

\begin{figure}[bp!]
    \centering
    \includegraphics[width=0.95\columnwidth]{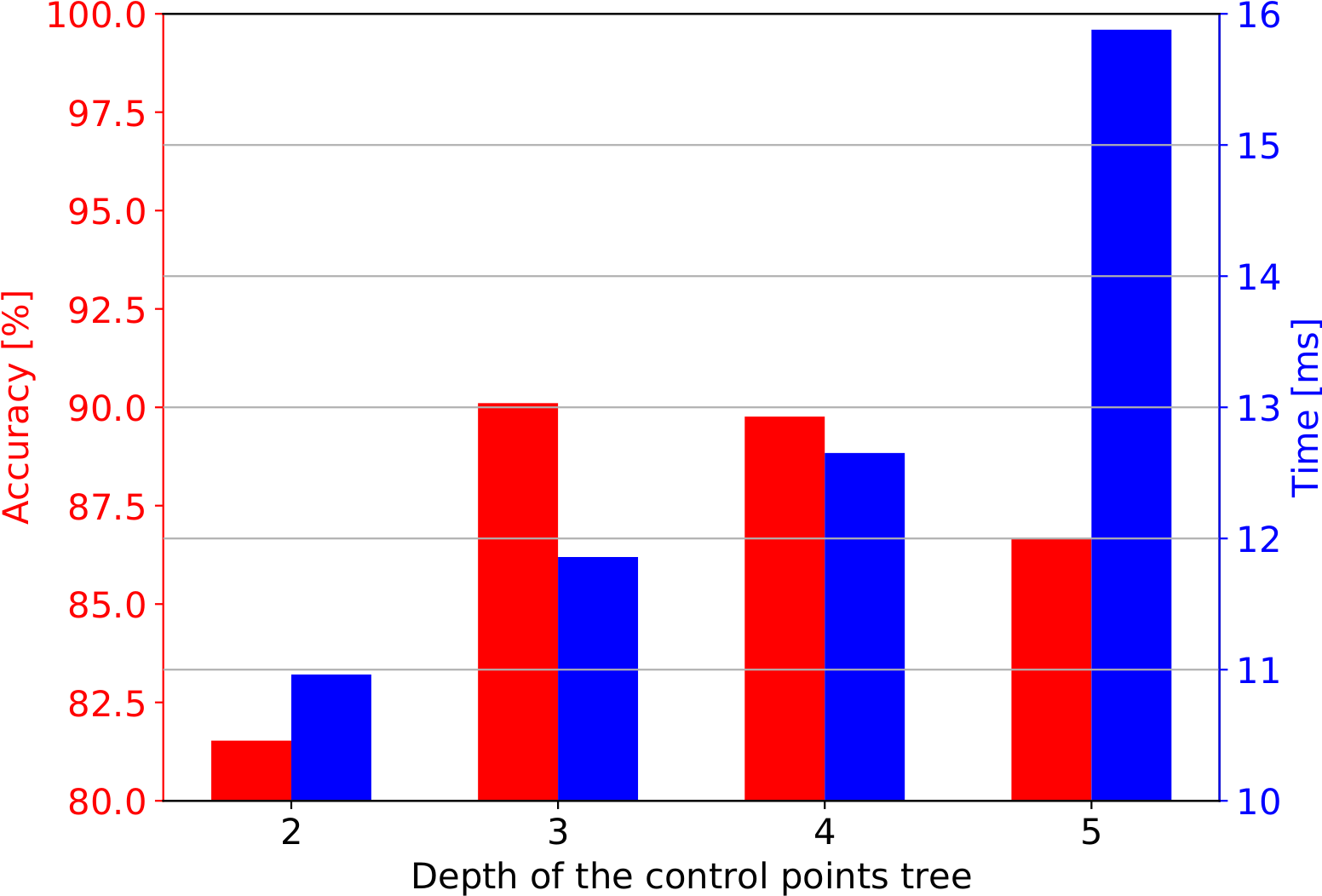}
    \caption{Accuracy and inference time with respect to the depth of the control points tree $D$. Model with $D=3$ achieves the best accuracy with reasonable low computational effort}
    \label{fig:N_dep}
\end{figure}

Firstly, we analyzed the dependency of the neural network-based motion planner from the depth of the control points tree $D$. 
Results of this experiment are presented in Figure~\ref{fig:N_dep}. One can see, that by increasing the depth of the tree, the running time grows, mainly because there is more time needed to interpret the neural network predictions as the number of control points grows exponentially with $D$. However, in the case of the accuracy of the models (understood as the ratio of feasible plans generated by the neural network to the overall number of the problems in the test set), the dependency is much less trivial. 
Accuracy, as expected, grows with the number of control points but only till the $D=3$, after which it decreases slightly. This phenomenon can be explained by the fixed capacity of the neural network, which may be not able to determine so many control points, and the fact that for problems that require a relatively short solution path it is hard to fit all control points, such that the resultant path does not violate the curvature constraints. Taking into account obtained results, we decided to use a neural network model with $D=3$ for further experiments.

\begin{figure}[tbp!]
    \centering
    \includegraphics[width=0.9\columnwidth]{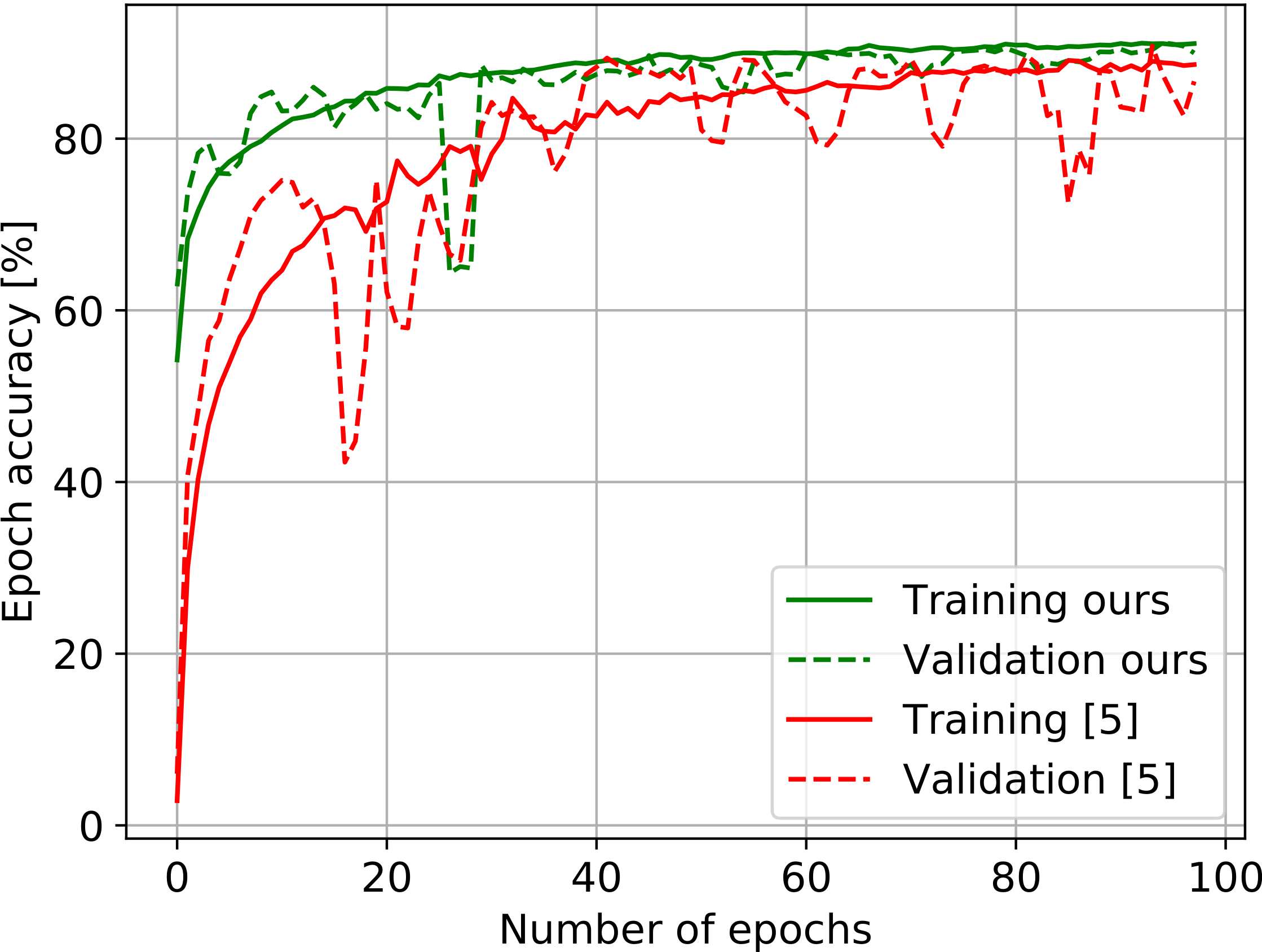}
    \caption{Accuracy on the training and validation sets obtained by the neural 
network planner proposed in this paper and in \cite{eaai2021} throughout the 
learning process}
    \label{fig:learning_curves}
\end{figure}

\begin{figure*}[htbp!]
    \centering    
    \includegraphics[width=\textwidth]{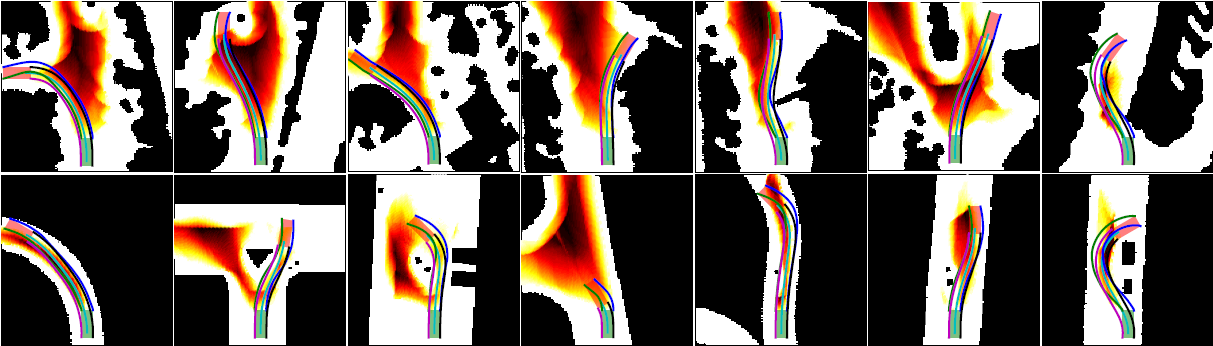}
    \vspace{-4mm}
    \caption{Visualizations of the reachable sets (presented as heatmaps: darker areas are reachable with greater orientations ranges) and sample paths generated by the proposed 
B-spline neural network motion planner on the maps unseen in the training process}
    \label{fig:as}
    \vspace{-1mm}
\end{figure*}

Then, we analyzed the learning curve of the proposed model for $D=3$ in comparison with the model from \cite{eaai2021}. In Figure \ref{fig:learning_curves} one can see the revealed inductive bias of the introduced representation, which enables the proposed neural network to achieve about $60\%$ accuracy on the validation set
just after the first epoch of training and about $80\%$ after the 10th epoch.


Figure~\ref{fig:as} shows exemplary paths generated with the use of the proposed B-spline neural network planner for the scenarios not included in the training set. Moreover, we present the generalization abilities of the proposed solution, by visualizing the sets of final configurations for which our model was able to 
generate a feasible path.

Table~\ref{tab:results} presents a numerical comparison of the proposed B-spline 
motion planning network with other path planning methods for a car-like vehicle 
with curvature constraints. 
This comparison was done on the test part of the dataset introduced in \cite{eaai2021}, with the use of planners from the motion planning benchmark Bench-MR \cite{benchmr} adapted to interface with the motion planning problems from \cite{eaai2021}.
This adoption required only to implement an interface to properly import the grid maps from our dataset and to make minor changes in the vehicle model. 
We believe that this approach, together with the open-source code of our planner, makes the presented results easily repeatable, and allows others to compare with our proposal. 

The first method which we considered was the State Lattice algorithm, 
implemented in the SBPL library \cite{SBPL}, which was operating on the grid with 
\SI{0.2}{\metre} spatial and $\frac{\pi}{16}$rad angular resolution. It was 
equipped with 11 different motion primitives per orientation, which were made 
with 3rd order B-splines with 0 curvature at the boundaries to preserve the 
continuity of the path curvature.
The second algorithm was the BIT$^*$ \cite{BIT} equipped with continuous curvature Dubins curves \cite{ccdubins}. We limited its running time to 
\SI{100}{\milli\second} to achieve a comparable time scale.
While we tested other sampling-based motion planning algorithms available in the Bench-MR framework, such as SORRT$^*$ \cite{ompl} and InformedRRT$^*$ \cite{informed}, their accuracy was about 20\% (even for maximal allowed planning time equal to \SI{1}{\second}), thus we chose only BIT* for the comparison.
Next baselines are the two versions of the neural motion planner introduced in
\cite{eaai2021}. 
The model directly taken from \cite{eaai2021} is denoted NMP I, whereas in the
case of the model named NMP II we reduced 10 times the allowed deviation between 
the desired and the actually reached configurations. 
Thus, NMP II shows how introducing more tight constraints on the final 
configuration impacts the performance.

\begin{figure}[htbp!]
    \centering
    \includegraphics[width=\linewidth]{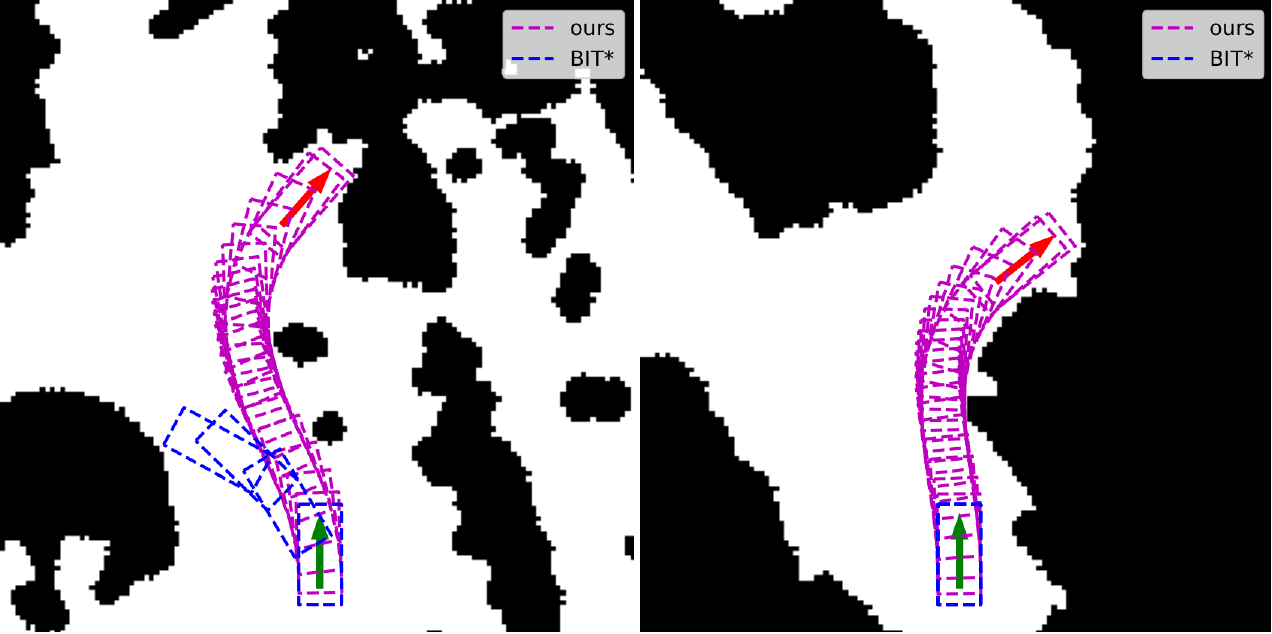}
    \vspace{-6mm}
    \caption{Paths generated by the proposed planner (purple) and BIT* with continuous curvature Dubins curves (blue) for two different motion planning problems which may be considered hard}
    \label{fig:hard}
\end{figure}

\begin{figure}[htbp!]
    \centering
    \includegraphics[width=0.95\linewidth]{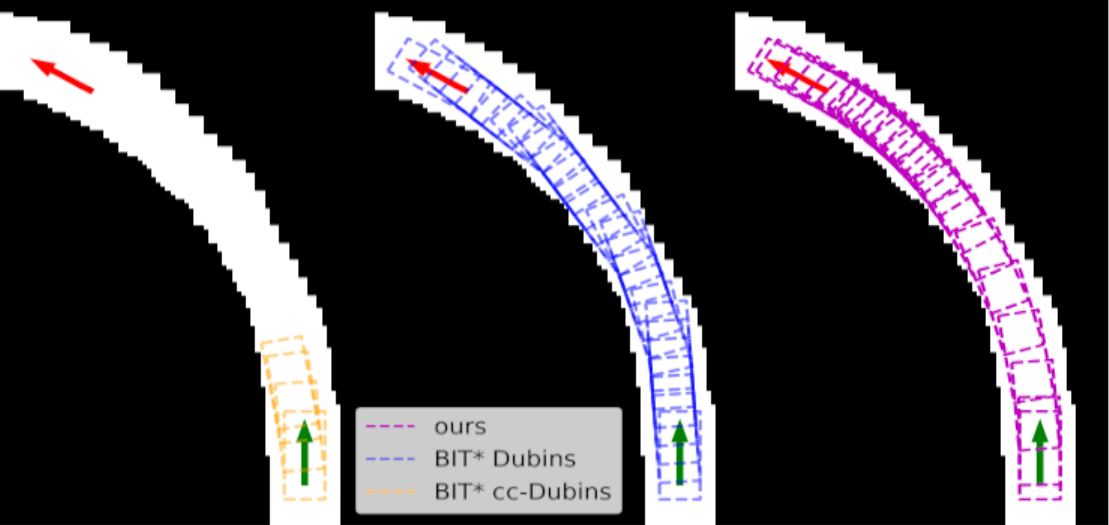}
    \vspace{-2mm}
    \caption{Paths  generated  by  the  proposed  planner  (purple)  and  BIT*  with continuous curvature Dubins curves (orange) and BIT* with plain Dubins curves (blue) for the turn scenario on the extremely narrow road}
    \label{fig:hard3}
    \vspace{-2mm}
\end{figure}

We can see that only the NMP I model achieved better accuracy than the 
one proposed in this paper.  
However, it is important to notice, that this model is allowed to finish the 
path in a subset of configurations around the goal, and as we can see in the 
case of NMP II, reduction of the size of this subset leads to degradation of the 
performance. Even lower accuracy was achieved by the SL and BIT$^*$ algorithms, 
which suggests that these methods are unable to plan in such a short time for 
more complicated scenarios.
Significantly better accuracy was achieved by the B-spline neural network 
planner (about 90\%), which also noted the lowest mean planning time equal to \SI{11}{\milli\second} (on i7 CPU)
that is almost constant for all test scenarios. 
What is more, the proposed planning method produces paths that are the smoothest ones, have the lowest mean of the maximal curvature ($\kappa$), and are free from
some tedious constraints such as zeroing the curvature at the ends of every segment
(SL) or generating paths with high maximal curvature (BIT*). 

\begin{table}[!th]
\centering
\caption{Accuracy, planning time, average maximal curvature of the path, and its 
order of continuity for following planning methods: state lattices \cite{SBPL,
benchmr} (SL), BIT* with continuous curvature Dubins curves \cite{BIT, benchmr}, 
neural motion planner \cite{eaai2021} (NMP I/II) and the solution proposed in 
this paper (ours).}
\begin{tabular}{c|ccccc}
\hline
Planner & SL & BIT* & NMP I & NMP II & ours\\
\hline
Accuracy [\%] & 49.4 & 72.83 & \textbf{92.24} & 79.25 & 90.1\\
Time [ms]& 23$\pm$27& 45$\pm$35 & 43$\pm$2 & 43$\pm$2 & \textbf{11$\pm$1}\\
Mean max $\kappa$ [$m^{-1}$] & 0.179 & 0.226 & 0.152 & 0.192 & \textbf{0.145} \\
Continuity & $\mathbb{G}^2$ & $\mathbb{G}^2$ & $\mathbb{G}^2$ & $\mathbb{G}^2$ & 
\textbf{$\mathbb{G}^5$}\\
\hline
\end{tabular}
\label{tab:results}
\end{table}

In Figures \ref{fig:hard} and \ref{fig:hard3} we present the performance of the proposed planner in comparison to BIT* (with Dubins and cc-Dubins extenders) on some relatively hard motion planning problems, each of which requires precise maneuvers in a confined space. For all considered problems our planner was able to plan a collision-free path, that satisfies the curvature constraints within \SI{12}{\milli\second}, in contrast to the BIT* with continuous curvature Dubins paths, that was unable to do so, even when run for \SI{100}{\second}. The reason it performs so poorly may be due to problems with sampling states that are possible to connect with cc-Dubins curves, and the need of including the clothoid segments between the straight lines and arcs, which is hard to perform when there is not enough space. To validate if resigning from the path curvature continuity will ease the planning process, we checked the BIT* algorithm with typical Dubins paths on a narrow turn. Results of this comparison are shown in Figure \ref{fig:hard3}. One can see, that using cc-Dubins paths makes it impossible to complete the maneuver, whereas plain Dubins curves reach the goal state, but collide with the environment in the narrowest part of the maneuver. This type of maneuver exposes the problems in planning with the use of curves which enables to plan fast, but at the same time imposes severe constraints on the curvature of path segments. In contrast, the proposed solution is able to plan extremely fast, while being able to produce smooth paths with a curvature tailored to the specific task.

\section{CONCLUSIONS}
\label{sec:conclusion}
We presented an important improvement to the DNN-based approach to local maneuver
generation for self-driving cars introduced in \cite{eaai2021},
that reduces the planning time to about \SI{11}{\milli\second},
ensures that the generated path reaches the desired configuration accurately,
and increases the order of path continuity.
The new architecture of our DNN introduces also an inductive bias due to the way the resultant path is constructed.
As result, the neural network is trained much faster than the previous one and achieves about 60\% accuracy just after the 1st epoch.

The conducted evaluation shows that the proposed method achieves 90.1\% accuracy 
on the test set outperforming other planning algorithms such as State Lattices
and BIT*, as well as the neural motion planner from \cite{eaai2021} for the reduced set of allowed final configurations. Moreover, our novel algorithm 
generates paths with the lowest mean of maximal curvatures for the
considered scenarios, which allows for higher velocities or smaller centrifugal forces.
The proposed B-spline neural motion planner generalizes to the previously unseen maps and scenarios and can handle typical local maneuvers, such as parking, sharp turns, avoiding obstacles, or navigating in narrow passages.
We have demonstrated in \cite{eaai2021} by CARLA simulations that our neural planner can handle dynamic environments.
The new approach should improve also on this aspect, due to the even shorter planning time.
However, considering dynamic environments directly by prediction of the occupied areas from the time sequence of local maps is in our agenda for future work.
Moreover, future work concerns the possibility to use our path generator as an initial 
guess for optimization-based motion planners, which can help to achieve higher accuracy in an acceptable time.
We are also working on the integration of the new planner with the CARLA simulation environment, as we did in the previous version, which, however, requires a substantial change in the path following controller.








\bibliographystyle{IEEEtran}
\bibliography{IEEEabrv,icra}

\end{document}